\def\year{2022}\relax
\documentclass[letterpaper]{article} % DO NOT CHANGE THIS
\usepackage{aaai22}  % DO NOT CHANGE THIS
\usepackage{times}  % DO NOT CHANGE THIS
\usepackage{helvet}  % DO NOT CHANGE THIS
\usepackage{courier}  % DO NOT CHANGE THIS
\usepackage[hyphens]{url}  % DO NOT CHANGE THIS
\usepackage{graphicx} % DO NOT CHANGE THIS
\usepackage{natbib}  % DO NOT CHANGE THIS AND DO NOT ADD ANY OPTIONS TO IT
\usepackage{caption} % DO NOT CHANGE THIS AND DO NOT ADD ANY OPTIONS TO IT
\DeclareCaptionStyle{ruled}{labelfont=normalfont,labelsep=colon,strut=off} % DO NOT CHANGE THIS
\usepackage{algorithm}
\usepackage{algorithmic}
\usepackage{enumitem}

\usepackage{newfloat}
\usepackage{listings}
\floatstyle{ruled}
\newfloat{listing}{tb}{lst}{}
\floatname{listing}{Listing}
\title{Braid: Weaving Symbolic and Neural Knowledge into 
Coherent Logical Explanations}
\author {
    Aditya Kalyanpur,
    Thomas Breloff, 
    David Ferrucci
}
\affiliations {
    Elemental Cognition Inc.\\
    \{adityak, tomb, davef\} @ec.ai
}

\usepackage{bibentry}
\begin{document}

\maketitle

\begin{abstract}
\noindent Traditional symbolic reasoning engines, while attractive for their precision and explicability, have a few major drawbacks: the use of brittle inference procedures that rely on exact matching (unification) of logical terms, an inability to deal with uncertainty, and the need for a precompiled rule-base of knowledge (the “knowledge acquisition” problem). To address these issues, we devise a novel logical reasoner called Braid, that supports probabilistic rules, and uses the notion of custom unification functions and dynamic rule generation to overcome the brittle matching and knowledge-gap problem prevalent in traditional reasoners. In this paper, we describe the reasoning algorithms used in Braid, and their implementation in a distributed task-based framework that builds proof/explanation graphs for an input query. We use a simple QA example from a children’s story to motivate Braid’s design and explain how the various components work together to produce a coherent logical explanation. Finally, we evaluate Braid on the ROC Story Cloze test and achieve close to state-of-the-art results while providing frame-based explanations.
\end{abstract}

\section{Introduction and Related Work}
KR\&R systems work well for certain knowledge-rich domains that typically involve a (pre-defined) set of axioms or rules, use structured queries and datasets, and have a need for precise logical inference with explanations. %Formal logic-based reasoning engines such as Cyc \cite{Lenat1985} and Ergo \cite{Grosof2018} have been successfully deployed in domains such as legal, healthcare and finance.  One of the main advantages of using such systems is transparency – the underlying reasoning of the system is well-understood and can be justified to end-users. 
However, they have several well-known limitations. The inference procedures are highly brittle in that they require precise matching/unification of logical terms to construct a complete explanation. Moreover, such systems suffer from the knowledge acquisition problem (i.e. how does one acquire the rules). Often, the rules are hand-coded, an approach which doesn’t scale in general. 

To address these issues, there has been a growing interest in exploring neuro-symbolic approaches, such as NL-Prolog \cite{Weber2019} in which the authors use a Prolog-like system to do back-chaining from a query to find proofs, where the entities/predicates have distributed representations (allowing for weak unification), and the inference rules are learned during training by specializing generic templates like \emph{P1(?X1, ?X3) :- P2(?X1, ?X2), P3(?X2, ?X3)}. However, NL-Prolog is an end-to-end (E2E) differentiable system whose explanations are not fully transparent (e.g. the learned rule predicates have distributed representations and are not directly interpretable). Also, the embeddings for entities/predicates once learned during training are fixed at test time. Other neuro-symbolic approaches are built on similar ideas to encode logical semantics in E2E differentiable networks such as Logic Tensor Networks \cite{Serafini2018} and IBM's Logic NNs \cite{DBLP:IBM_LNN}. 

We take a complementary approach: our novel logical reasoner, called Braid, is at its core a symbolic reasoning engine, and thus capable of producing explicit logical explanations, but at the same time, it uses statistical methods to \emph{inject} explicit term alignment or missing rule knowledge, considering the Knowledge Base (KB) context at run-time.

It does this by supporting \emph{custom-unifiers}, which are functions that propose and score mappings between the terms of two logical propositions, given the KB as context. In our work, we use neural matching functions as unifiers. %Their purpose is to help the reasoner find proofs even when (sub) goals, rule conditions and/or facts do not align perfectly. 
Braid also supports \emph{dynamic rule-generators} - DRGs - which given a target proposition (goal) and a KB as input, output a scored list of hypothesized rules that are used to prove the target (these rules are fully interpretable, unlike in NL-Prolog). We describe a DRG implementation using a neural rule generation model that was trained on a dataset of crowd-sourced causal rules \cite{Glucose}.
   
We discuss the reasoning algorithms used in Braid, their implementation in a distributed task-based framework for proof graph building, and an evaluation of Braid on ROC stories \cite{ROC}.

%Our approach shares some similarities with the RETE framework \cite{RETE} for matching production rules (e.g. reusing nodes for the same sub-goal proposition, message passing between nodes etc.) but makes several novel extensions: we primarily do backward chaining via a heuristic best-first search (A* search), leverage a Master-Worker architecture where the Master builds the main proof graph while Workers make local inferential updates, and define general functions for Unifiers and Provers that lets us plug in various reasoning strategies combining standard reasoning (e.g. syntactic resolution based) with statistical (e.g. word-embedding based) approaches.

\section{Motivating Example}
\label{section:motivation}
We describe an example to illustrate the challenges that a logical reasoning engine faces when answering story understanding questions. The  text below is from a children’s story (Grade: K) on the ReadWorks website (http://readworks.org).

Consider the following short story: 

\emph{“Fernando and Zoey go to a plant sale. They buy mint plants. They like the minty smell of leaves. Zoey puts her plant near a sunny window. The plant looks green and healthy!”}

The question we would like to answer is: 

\emph{“Why does Zoey place the plant near the window?”}

Questions of this nature are part of the \textbf{Template of Understanding}, defined in \cite{jesse2020}, that is used to test an AI system's deep understanding of a narrative story. The paper demonstrates that existing SOTA question-answering systems perform very poorly at this task, which remains an unsolved research problem.

\begin{figure}[h]
\centering
\includegraphics[width=2.5in]{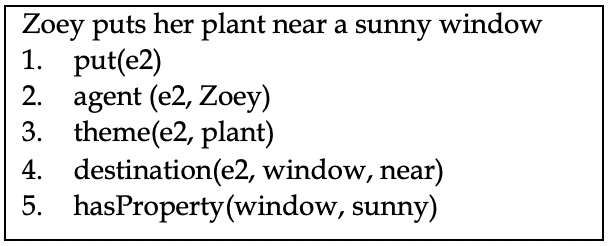}
\caption{Logical interpretation of a story sentence}
\label{fig:story-interp}
\end{figure}

Figure \ref{fig:story-interp} shows the logical interpretation of a sentence in the story. For each sentence, we generate multiple probabilistic interpretations (each associated with a confidence) by running a state-of-the-art semantic parser \cite{spindle2020} and a co-reference resolution component (Stanford). We omit details about the NLP stack, as they are not relevant for this paper.
   
 In the figure, \emph{e2} is a constant which denotes an event/action, while \emph{Zoey, plant} etc. are constants derived from story terms. On the other hand, predicates such as \emph{put, agent, theme} etc. come from our lexical ontology called Hector. The Hector ontology is a collection of frames (concepts and relations) derived from FrameNet \cite{framenet:book} and NOAD \cite{noad} that aims to capture the core meaning behind text. 

\begin{figure}[h]
\centering
\includegraphics[width=3in]{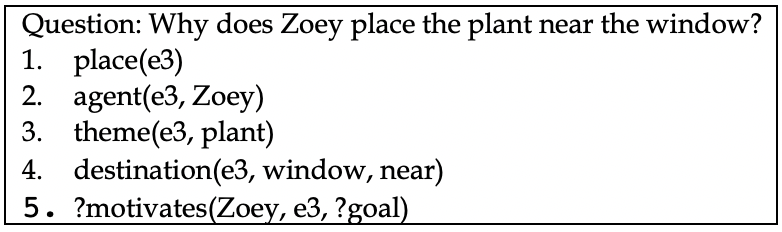}
\caption{Logical interpretation of question}
\label{fig:question-interp}
\end{figure}

The question interpretation is shown in Figure \ref{fig:question-interp}. We assume that we have run a co-reference algorithm on the question and story text, and so the same constants are used (when co-referential) in the question and story interpretations. The specific question representation (line 5) uses the \emph{motivates} relation and is querying for Zoey’s goal which explains her performing the place action. 
 
Answering this seemingly straightforward question requires a lot of implicit background knowledge, such as that a plant near a sunny window gets exposed to light, that plants need light to be healthy, and that Zoey wants the plant to stay alive (which motivated her action). Also, in this particular example, the \emph{put} action (in the story) and the \emph{place} action (in the question) are similar/synonymous in this context.
   
 In the subsequent sections, we shall see how Braid resolves these issues via dynamic rule generation (to bring in background knowledge) and fuzzy unification (to overcome the verb action mismatch). 

\begin{figure*}
\centering
\includegraphics[width=5.5in]{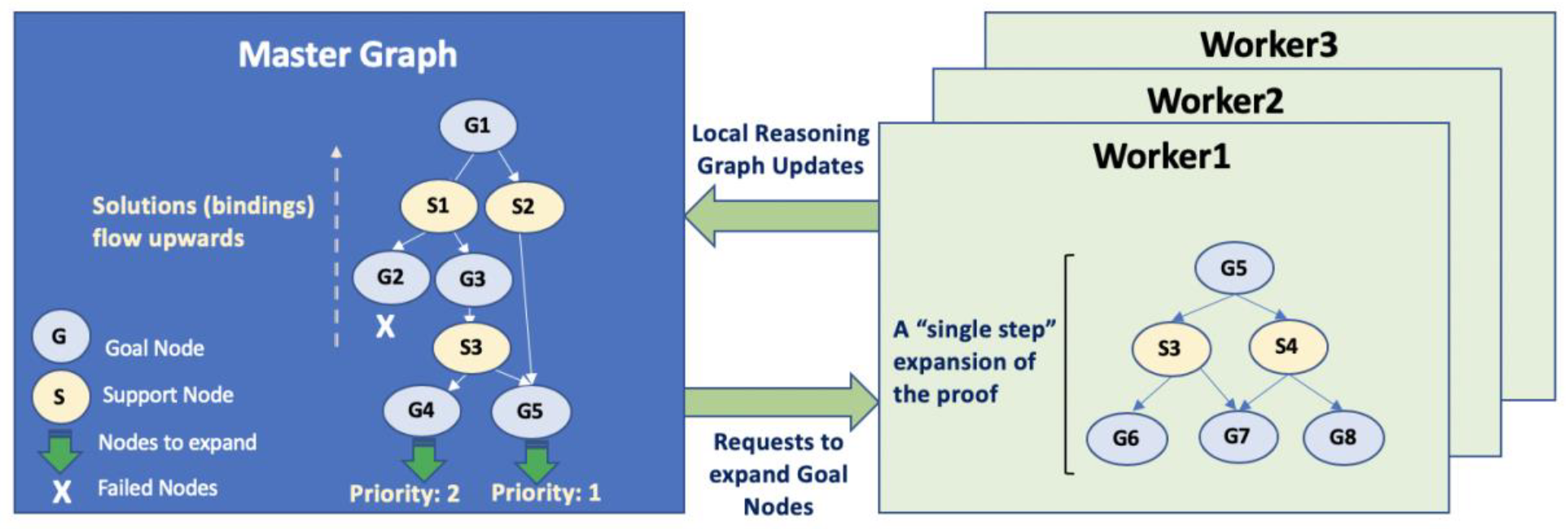}
\caption{Distributed Proof Graph Builder}
\label{fig:arch}
\end{figure*}

\section{Braid}
\label{section:braidIntro}

In this section, we describe the overall Braid framework – a parallelized infrastructure for constructing deductive proof/explanation graphs for a given query and KB. %The framework leverages two core logical functions which we discuss first – \emph{Unifiers} and \emph{Provers}, before we delve into the larger reasoning architecture.
We start with two key Braid functions: Unifiers and Provers.

\subsection{Unifiers}
\label{sec:unifiers}

 One of the core functions in any FOL-based reasoner is unification. The standard unification function (\emph{syntactic unification}) takes a pair of predicate logic formulae P1, P2, and checks if there exists a mapping of variables from one to the other which makes the two formulae equal. For example, the formulae \emph{hasPossession(Zoey, ?y)} and \emph{hasPossession(?x, plant)} unify with the mapping \emph{[?x=Zoey, ?y=plant]}.
 
 We generalize the notion of unification to be any FOL formulae matching function, defined as follows: 

 \setlength{\parindent}{2ex} \texttt{unify(P1, P2, K) $\rightarrow$ \{UR1...URn\}}

 \noindent where P1, P2 are predicate logic formulae, K is the knowledge base (which acts as context for the matching) and \{UR1..URn\} are a set of \textbf{Unification Results}, where each UR contains the following information: a substitution that maps variables, entities \emph{or even the predicate} in one formula to the other; score (0-1) which reflects the confidence of the mapping; and additional metadata used by the unifier function when making its decision (which can be exposed by Braid in the final explanation). 
   
 Consider the example described in the previous section, where the story interpretation contains the proposition \emph{put(e1)} while the question interpretation has the proposition \emph{place(e3)}. Under standard unification, both these propositions would not unify, as they use different predicates and arguments. However, we have designed a custom unification function that considers word/phrase similarity (note: \emph{place} and \emph{put} are constants derived from story text) to align the two formulae, using the additional context that they share the same agent (\emph{Zoey}), theme (\emph{plant}) and destination (i.e. \emph{near the window}) to boost the match score. Such a function may return \texttt{unify(put(e1), place(e2)) $\rightarrow$ (\{put=place, e1=e2\}, 0.9)}.
 
Our default algorithm for fuzzy unification between (P1, P2) uses BERT to generate distributed representations for the predicates and arguments in the two formulae respectively, by running on the text the symbols were parsed from, then gets the pair-wise cosine-sim across the terms, and outputs a weighted sum of the similarity scores. If P1 and P2 are reified events, the algorithm computes a product of the scores for all the event-participant relations.

\subsection{Provers}

Similar to how SLD resolution (Prolog) works by building out a search tree for a given query/goal, Braid works by constructing a proof graph by using unification methods to backchain on clauses (rules/facts) in the KB. 
   
To support various reasoning algorithms, we define the notion of a Prover, a function which given a (sub) goal and the KB, performs a “single step” expansion of the graph along a particular reasoning path. 
  
\setlength{\parindent}{4ex}  	 \texttt{prover(G, K)  $\rightarrow$ PD}

\noindent where G is the input goal, K the knowledge base, and PD is a partial proof-derivation graph with the following properties:

\begin{itemize}[noitemsep]
    \item PD has two types of nodes: goal nodes and support nodes. 
    \item Support nodes provide justification for some goal node and play a key role in the flow of information (e.g. solution bindings) in the overall proof graph. 
	\item Goal and support nodes in the graph are interleaved, i.e. a goal node can only have support nodes as its children (and vice versa).
    \item G (input goal node) has to be the root of the graph PD
\end{itemize}

For example, a \emph{Rule-Based Prover} finds rules in the input KB whose respective consequents unify with the goal (using any implementation of the Unification interface defined earlier), and then output a partial proof-tree which has the goal as its root, a \texttt{Rule-Support} child node for each such rule whose consequents unify, and an additional edge from each support node to the corresponding antecedent of the rule. Note that the prover need not prove the antecedent immediately. 
   
The main advantage of this design is its scalability for deployment in a distributed setting – each prover performs a local computation, without requiring knowledge about the overall proof-graph, which enables parallelization across cores and machines. Also, the communication between the master Braid algorithm (which constructs the entire proof graph) and each individual prover is kept to a minimum.

\subsection{Distributed Proof Graph Builder}

Braid uses a task-based framework where a central “Master” task builds the entire proof graph for the input goal by communicating with a set of “Worker” tasks, each of which use \texttt{Provers} to perform local graph-based reasoning (see Figure \ref{fig:arch} for the architecture diagram).

The master algorithm is a generic, parallelized, graph building approach that continuously modifies a central Braid graph based on asynchronous updates coming in from (remote) workers.

Nodes in the Braid graph are either Goal nodes or Support nodes. Each node is associated with a collection of Unification-Results that represent bindings flowing into the node via its child edges. Crucially, atomic goal nodes are reused across the graph (i.e. there is only one node per atomic goal proposition in the entire graph) which allows us to reuse solutions found earlier for the same goal proposition (caching). 

Goal nodes have a State, which is one of three values: \texttt{Success, Failure and Unknown}. A special kind of support node is a \texttt{Fact-Support} node, which are graph leafs, and correspond to facts that unify with atomic goals. %Each support node is associated with a Prover which specifies the logic for handling new binding updates coming into that node. %Finally, since adjoining nodes may have propositions involving different variables, edges containing nodes are associated with variable-variable mappings.   

The key functions of the master algorithm are:

\begin{itemize}[noitemsep]
\item \textit{Merging Local Graph changes into the central Master Graph}: which takes the next graph update coming from the worker and merges its contents into the main graph.  %The event might be ignored (if we have already merged an equivalent event) or it may be changed to maintain internal consistency.  
If we are merging a new edge from the local graph sent by the worker, and there is an equivalent (under variable substitution) destination goal node in the master graph, we connect the new local goal node to a new \texttt{Variable-Mapping Support} node (which stores the underlying variable substitution) and connect the latter to the corresponding master goal node. 
\item \textit{Determining Next Sub Goal}: uses a Goal Selection Strategy to determine which goal node(s) in the graph to expand next. %Goal nodes initially start off in the \texttt{Unknown} state. When picked by the strategy, they are moved to the \texttt{Failure} state and they can later switch to the \texttt{Success} state when solution bindings are propagated to it. When a worker becomes available, the strategy selects the next-best goal to expand and sends it to the worker. 
%Cycles are automatically blocked early since each master goal node is expanded at-most once. 
The default ordering strategy uses the following features: an aggregation of the product of confidences of the supports in the path from the node to a query, the minimum path distance from the node to a query, a measure of goal complexity (computed using syntactic heuristics such as the number of nested formula in the goal), and a measure of goal “plausibility” (estimated by using a statistical language model to score text plausibility on the textual version of the goal proposition). 
\end{itemize}

Detailed pseudo-code of the proof-graph builder algorithms are in the supplementary material (Appendix).

\section{Reasoning Algorithms in Braid}
   
We now describe two reasoning algorithms that have been implemented as Provers in Braid: a default reasoner that extends SLD and can be used to explain any query, and a specialized reasoner for explaining why agents carry out certain actions in a given story.

\subsection{Default Backchaining Prover: SLD+}
\label{ref:sec:SLD}

\begin{figure}[h]
\centering
\fbox{\includegraphics[width=3.5in]{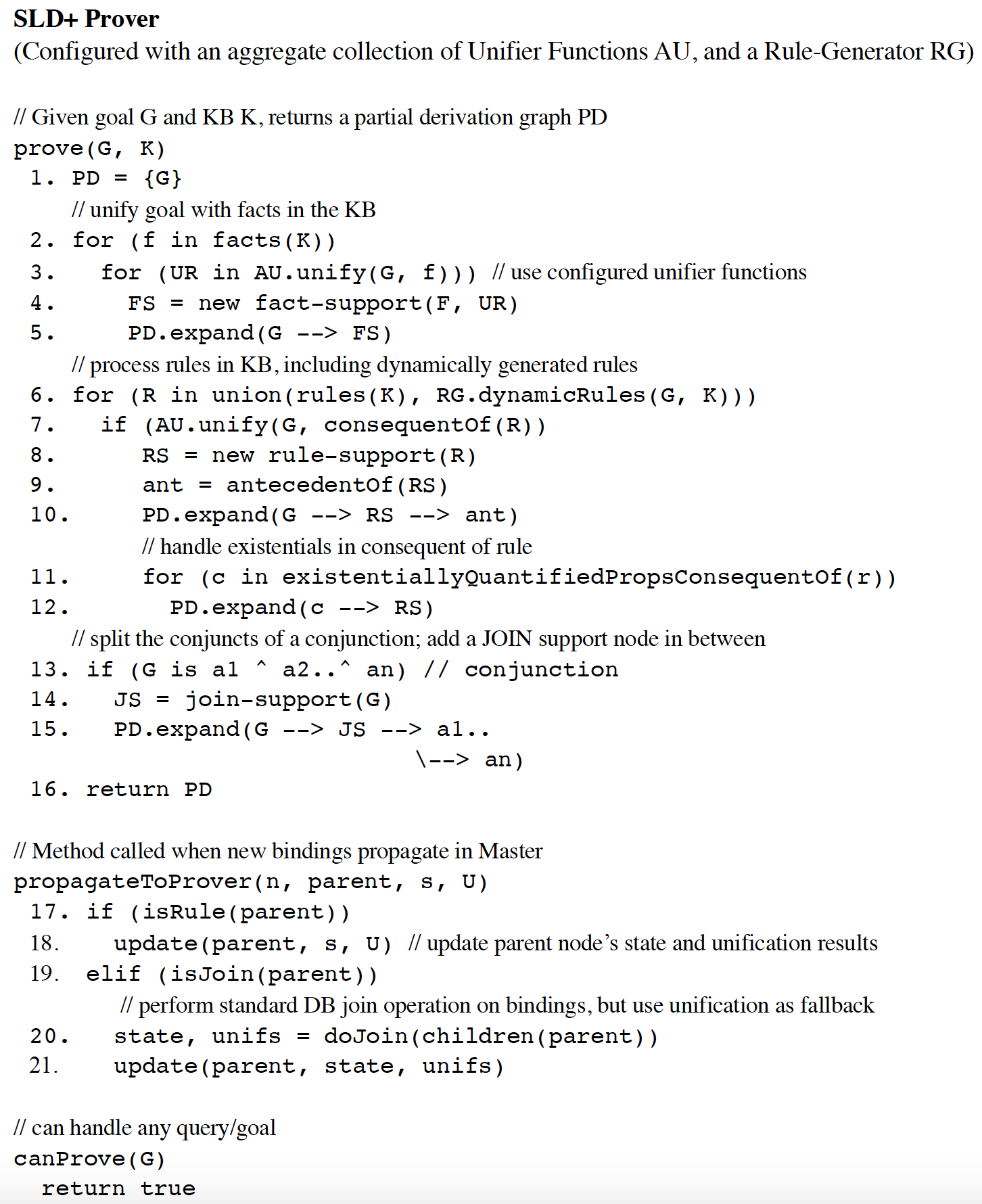}}
\caption{Default Prover based on SLD resolution with extensions}
\label{fig:sld}
\end{figure}

The default back-chainer in Braid is based on SLD resolution with a few key extensions (hence the name SLD+). The algorithm is shown in Figure \ref{fig:sld}.

Like SLD resolution, it tries to resolve the goal by unifying it with clauses (facts/rules) in the KB. Unlike standard SLD, however, it uses custom unifier functions when performing the unification (/matching) in three separate places: when matching the goal against facts in the KB (line 3), when unifying against the consequents of rules (line 7), and in the join operation when bindings are propagated (line 18) as a fallback if the standard DB join operation fails. As noted earlier, the unifiers return Unification-Results with a matching score which is used by Braid to score and rank the solutions and proofs for the input query.
   
Additional differences from standard implementations of SLD resolution include: the algorithm splits conjunctive goals into individual conjuncts which can be evaluated in parallel (lines 13-15); it uses a Rule-generator to dynamically generate rules for the given goal (line 6); and it creates additional nodes for rule consequents containing existentially quantified variables (lines 11-12), in order to forward propagate skolemized inferences.

\subsection{“Agentful” Action Prover}
\label{section:AAP}

%\begin{figure}[h]
%\centering
%\fbox{\includegraphics[width=3.5in]{AAProver2.png}}
%\caption{Prover to explain the rationale for Agentful action}
%\label{fig:aap}
%\end{figure}

For the motivating example, we need to explain the action of Zoey placing her plant near the sunny window. In general, we refer to such actions as Agentful actions (the agent being Zoey in this case). 
   
One reasoning strategy for explaining Agentful actions is the following: first find the motivations of the agent, and then check if the action carried out by the agent leads to one of the agent’s objective.
   
This reasoning strategy can be described by the following logical rule:

\texttt{motivates(?agent, ?action, ?goal)  :- hasGoal(?agent, ?goal), leadsTo(?action, ?goal)}

The rule uses the predicate \texttt{leadsTo} which has the following operational semantics: find a proof for the goal and check if the proof contains the specified action.

The default SLD+ prover does not have support for handling specialized predicates like \texttt{leadsTo}. Moreover, for efficiency sake, the antecedents of the rule should be evaluated in the order specified above – i.e. first, find the goals of the agent, and then for each goal, look for a proof for that goal and check if the action is part of the proof-tree. We implement this single-rule logic using a specialized prover called the Agentful Action Prover. In this prover implementation, when bound agent-goal solutions (for the first antecedent) are propagated back to the prover support node, it searches for proofs for the second (now fully bound) antecedent, and finally checks if the action is part of any proof. %Specifying a particular ordering for 

\section{Dynamic Rule Generator (DRG)}
\label{section:DRG}

As noted in the introduction, the idea behind dynamic rule-generation is to provide missing rule knowledge on the fly to the reasoning engine, where the rules can come from an external function (the rule generator). The generic DRG interface has one core method, which given a target goal and KB, returns rules relevant to proving the goal. We now describe a neural DRG implementation.

\begin{figure}[h]
\centering
\fbox{\includegraphics[width=3.5in]{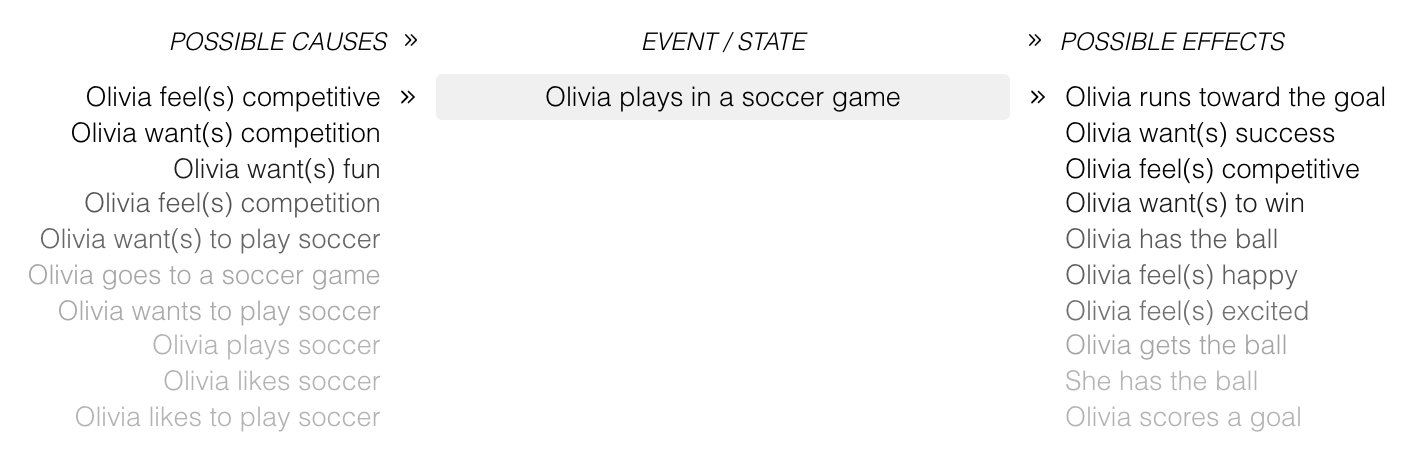}}
\caption{GLUCOSE suggestion for a story about soccer.}
\label{fig:glucose}
\end{figure}

GLUCOSE is a crowd-sourced dataset of around 500K common-sense explanatory rules. The dataset consists of both general and specific semi-structured inference rules that apply to short children's stories. The rules are collected along ten dimensions of causal explanations, focusing on events, states, motivations,  emotions,  and naive psychology. 

\cite{Glucose} had shown that by fine-tuning pre-trained transformer models on the GLUCOSE data, the resultant neural generative model was able to produce contextual common sense inference rules on unseen stories with surprisingly high accuracy. Example inferences by the Glucose DRG on a children's story are shown in Figure \ref{fig:glucose}. 

We use the GLUCOSE trained model as our DRG implementation for the deep story understanding problem, since the GLUCOSE rules were acquired along the same dimensions defined in the Template of Understanding. In other applications (such as in the next section), we create alternate DRG implementations more suited to the given task. 

%Since Braid is a symbolic reasoning engine, and the GLUCOSE DRG produces unstructured textual rules, we need to convert them into a structured logical form, and for this we use our Spindle parser

The input to the GLUCOSE DRG is a goal proposition and the KB propositions as context, and we run our Natural Language Generation (NLG) component (a straightforward template-based approach) to produce textual versions of all the logical formulae. We then feed the textual goal and context to the neural GLUCOSE model which returns textual rules with confidences. Finally, we convert these unstructured rules back into a structured logical form using the semantic parser.%\footnote{The crowd-sourced rules were collected in a semi-structured form (using subject-verb-object templates), and hence the rule expression text is regularized, which makes the parsing task easier}. 

\section{Proof Finding and Ranking using ILP}
\label{sec:ilp}

Once the Braid Master has built the master proof-graph for an input query, which encapsulates a collection of proof trees, the final step is to extract a ranked list of valid proofs for the query answers, provided the status of the original goal (query) node is Success. %There could be multiple answers for a given query, and each of those answers could have multiple proof-trees. 

%Since rules used in Braid are associated with confidences, and some facts may be brought in via a fuzzy unification (with an associated confidence), we need to take these confidences into account when scoring proofs.% and find a proof that maximizes the overall confidence of the rules/facts used in it.

%\footnote{This step is customizable - see Section \ref{sec:RocEval} where we use an alternate algorithm for proof scoring.} 

The default proof-finding procedure in Braid is cast as an Integer Linear Programming (ILP) problem: we create binary variables that represent the inclusion/exclusion of each node in the graph. %As mentioned earlier, there are two types of nodes in the graph – goal nodes (e.g. original query, any inference from a rule, or fact in the KB) and support nodes linked to provers (e.g. rule). 

The ILP program has the following constraints
\begin{itemize}[noitemsep]
\item The query/solution goal node must be included
\item If a goal node is included, exactly one of its support nodes must be included
\item If a support node is included, all its antecedent (children) goal nodes must be included; unless it is a \texttt{Fact-Support} node, in which case it is a leaf node
\end{itemize}    
Along with the following objective function: maximize the product of confidences of all included nodes.
    
Each solution to the ILP problem constitutes a single proof-tree to the input query, and the top solution has the highest overall confidence. %Note that the objective function can be modified to consider other ways (besides product, which assumes independence) to aggregate the confidences of the individual rules/facts.

\section{Putting it all together}
   
\begin{figure*}[htb]
\centering
\fbox{\includegraphics[width=5.5in]{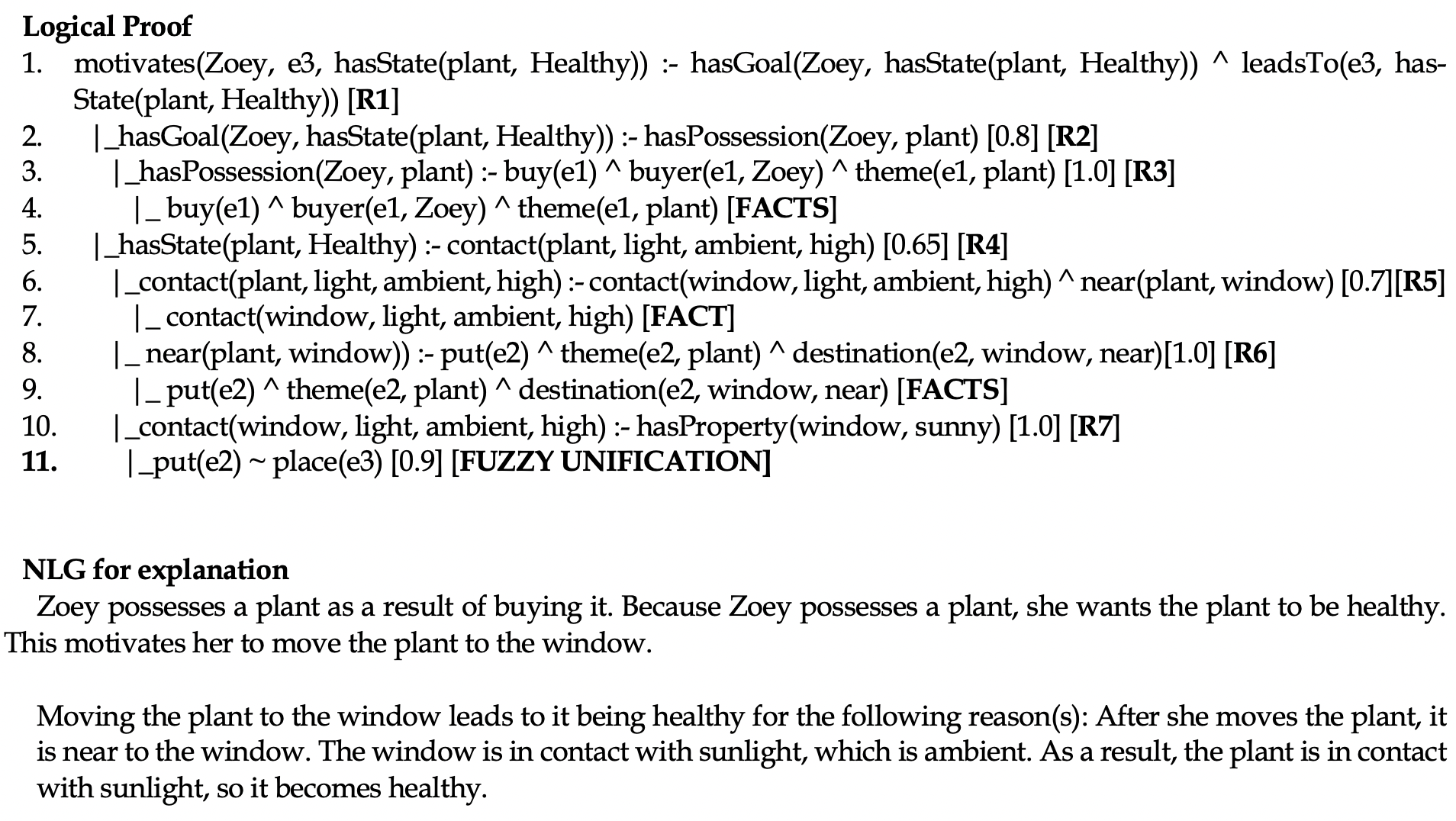}}
\caption{\textbf{Explanation for: Why does Zoey place her plant near the window?} %Rule R1 is the generic pattern captured in the Agentful-Action prover; R2, R4, R5 are dynamically generated; while R3, R6 and R7 fall out from the semantics of the underyling lexical ontology Hector.
Note that FACTS in the proof, which are generated by the NL interpretation components, are also associated with confidences (not shown above) that are factored into the final explanation score. Also, we have an Natural Language Generation (NLG) module, beyond the scope of this paper, which generates a more consumable explanation for end-users from the logical proof tree.}
\label{fig:explanation}
\end{figure*}

Braid solves the motivating example by generating the proof tree shown in Figure \ref{fig:explanation}. 
 
It starts by using the Agentful Action Prover (AAP) to backchain on the rule R1, which first looks for Zoey’s goals and then checks if the “place” action (e3) performed by Zoey can lead to one of her goals (line 1). The sub-proofs use rules to infer that Zoey wants her plant to be healthy (R2), and that the plant being in contact with light causes it to become healthy (R4). Note that R2, R4 and R5 are dynamically generated by the GLUCOSE-based DRG, while R3, R6 and R7 come from the semantics of “buy”, “put” and “sunny” respectively, as manually defined in the Hector ontology. Finally, a fuzzy unification function is used to realize that put and place are similar actions in this context.

While this example shows how Braid's components come together to produce a deep logical explanation in keeping with the Template of Understanding for narratives, more work is needed to demonstrate high accuracy across a large story set. However, preliminary results are promising - on a small sample of 10 children's stories from ReadWorks, Braid gets an accuracy of ~60\% over 50 questions in the "zero-shot" setting (i.e. without any training; since we have no ground truth data for explanations) which is substantially better than SOTA systems (38\%) at this task. Also note that with Braid the user can give feedback on each element of the proof (e.g. a faulty fuzzy unification or an invalid rule), and this knowledge is directly used as training to improve the underlying neural function. Getting this fine-grained training signal is not possible with other neuro-symbolic systems. 

%\section{Hints}

%When Braid fails to find an explanation for a query, it can generate relevant missing propositions, which if true, would help it complete the full proof​. We refer to such missing propositions as hints.
    
%While the search space of potential missing propositions is huge, we are interested in “near misses” – cases where the proof graph is close to completion. We use several features to score failure nodes in the proof graph – depth in the proof tree (from the original query), rule-percentage completion (if the node is one of the antecedents of a rule), parent rule confidence etc. For example, suppose we have a high-confidence rule that directly leads to the query, and the rule has five antecedent propositions, four of which are marked as successful and the last one is a failure, then the propositions associated with this failure node represents a near-miss, and makes for an interesting hint proposition (since if it were true, the rule would fire and directly conclude the goal, yielding a proof).

\section{Evaluation: ROC Story Cloze Test}
\label{sec:RocEval}

To demonstrate the effectiveness of using explicit semantic representations and dynamic rule generation for a more constrained NLP reasoning task (as compared to general deep story understanding), we used Braid to tackle the ROC Story Cloze Test. The Story Cloze task is as follows: given a 4 sentence story, and two possible endings, pick the more plausible story ending (one that fits better in the context of the story). Examples are shown in Table \ref{table:rocExamples}. For our experiments, we focus on the Spring 2016 ROC dataset, which has a validation set and a test set of 1871 examples each.

This task relies on commonsense knowledge, which is typically hard to acquire and represent explicitly. Nevertheless, recent advances in transformer-based language models, such as GPT2 and BERT, have made it possible to build E2E neural solutions that do very well (accuracy in the high 80s) on this task. A common downside with these approaches is the lack of explicability. Instead of a pure neural approach, we developed a hybrid neuro-symbolic solution to this problem using Braid, one that is capable of giving an explanation for choosing a particular story ending.

\begin{table}[htb]
    \centering
    \small
     \begin{tabular}{|p{4.5cm}|p{1.5cm}|p{1.25cm}|}
        \hline
        \textbf{Story} & \textbf{Right Ending} & \textbf{Wrong Ending} \\\hline
Rick grew up in a troubled household. He never found good support in family, and turned to gangs. It wasn't long before Rick got shot in a robbery. The incident caused him to turn a new leaf. & He is happy now & He joined a gang \\ 
        \hline
        Ignacio wants to play a sport while he is in college. Since he was a good swimmer, he decides to try out for swim the team. Ignacio makes it onto the team easily.	At the first swim meet, Ignacio wins second place! &	Ignacio gave up swimming. & Ignacio won a silver medal. \\ \hline
        Nya had been asked on a paintball trip with friends. She was nervous about going. But she went anyways, hoping to have fun. She shot paintballs at her friends and laughed the whole time. & She loved it so much she planned a trip for the next week. & She was shot and vowed to never go there again. \\ \hline
    \end{tabular}
    \caption{ROC Story Cloze Test Examples}
    \label{table:rocExamples}
\end{table}

Our approach is based on the notion of Schank's scripts \cite{scripts}, and assumes that each of the stories involves one or more frames or situations, and that information in the story is consistent with the frames. For example, the first row in Table \ref{table:rocExamples} might correspond to the frame \emph{lesson-learned} since Rick realizes the folly of his ways; the second to the frame \emph{made-plan-executed} as Ignacio has a goal in mind and works hard to achieve it; while the third row might correspond to the frame \emph{change-belief-happy} where Nya was initially nervous about a trip that she eventually enjoyed. 

Our hypothesis is that if we can correctly detect the applicable frame from the first 4 sentences of the story, we should be able to predict the right ending that is consistent with the detected frame. Furthermore, the frame provides an explanation for the chosen ending.

\subsection{Frame Inference}

\begin{figure*}[h]
\centering
\fbox{\includegraphics[width=6.5in]{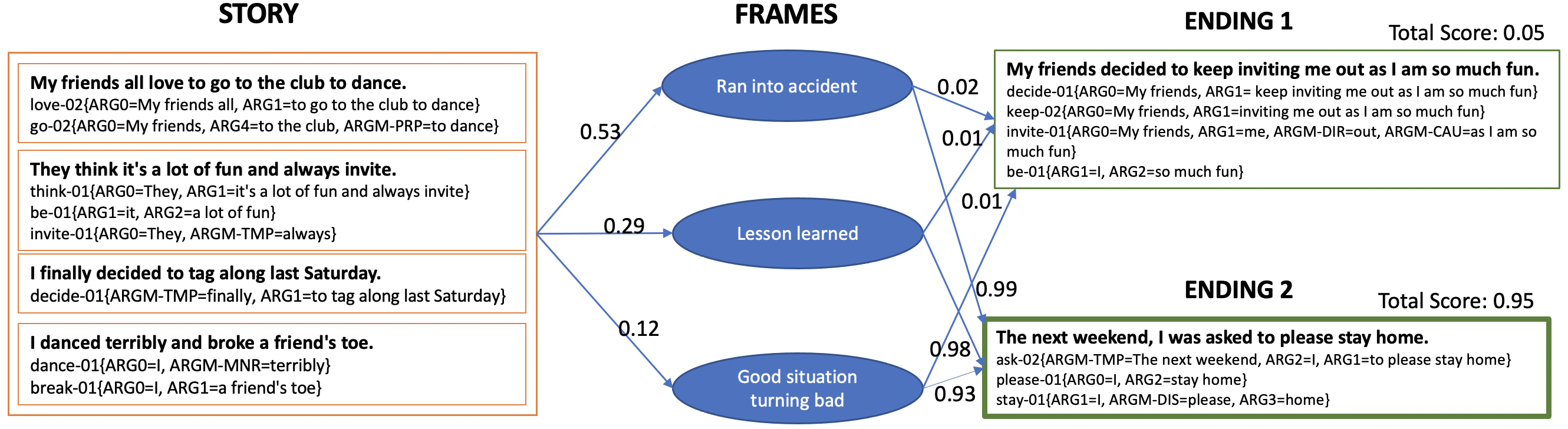}}
\caption{\textbf{Example of Frame-Based Explanation for a ROC Story Ending.} The figure shows the inference paths found by Braid using the neural classifiers (for detecting frames given story, and ending given frame+story), along with the corresponding rule scores. The semantic interpretations are shown below each sentence. Only the top-3 frames predicted for the story are shown in the figure (Note: Final scores consider the entire frame distribution).}
\label{fig:rocInference}
\end{figure*}

We decompose the Story Cloze problem into two steps: 
\begin{enumerate}
    \item Detect a frame given the first four sentences of the story, i.e. estimate $Pr(Frame|Story)$
    \item For each possible ending, compute the probability of it being the ending of the story considering the detected frame, i.e. $Pr(Ending|Story, Frame)$
\end{enumerate}

Our first challenge was to define a collection of relevant frames for the ROC stories. We analyzed a random sample of 350 story examples from the validation set (approximately 1/5th of the entire set) and manually labeled each story with a frame. Each frame is a generic description of the situation unfolding in the story. The process of frame creation was iterative but admittedly subjective - we generalized concepts that seemed too specific, or specialized highly generic concepts to capture sentiment/emotion. In all, we created a set of 19 frames. Examples include: \emph{met expectations (happy/sad)}, \emph{surprise (pleasant/unpleasant)}, \emph{resolving problem}, \emph{persistence (worked/failed)}, \emph{ran into accident} etc.

We used this annotated set to train a classifier using a pre-trained encoder-decoder model (T5-base), which given the first four sentences of the story as input, predicts the corresponding frame. We used a 90/10 split for train/dev and found that the classifier accuracy to be 87.5\% on the dev set. The high accuracy with relatively little training data is not surprising given the power of pre-trained transformers and the nature of this task. We then applied the trained frame detection classifier to the entire validation set in order to label the frames for all the 1871 stories in it. This gave us a ``bronze" dataset of frame labels for each story. 

The bronze frame labels were used to create larger training sets (from the entire validation set) for the two steps described above, i.e., classifying a frame given a story, and classifying whether a given ending to a story is consistent with the frame. Moreover, we created two different train sets - one using raw text, and the other using semantic parses, obtained by running the semantic parser on each of the story and ending sentences. The training data format for the second step was \emph{input}: \texttt{story SEP ending SEP frame}, \emph{output}: \texttt{1/0} for right/wrong story ending (SEP stands for a separator token). Finally, we built neural generative models using the above training data. %The model lets us compute a confidence (0-1) for a specific ending given the story context and frame. 

The models from the two steps are used in neural dynamic rule generators (DRGs): the first which produces rules of the form: \texttt{frame :- story}, while the second generates  \texttt{ending :- story, frame} rules. The rules have confidences coming from the corresponding classifier.

\subsection{Experiments}

%We ran the following four experiments:
\begin{enumerate}
    \item \textbf{Baseline}: Our baseline system is an E2E neural model. We train a binary classifier (fine-tuning T5-base) on the validation set to predict whether a given ending to a story is correct. At test time, we apply the classifier to both endings and pick the one with the higher score.
    \item \textbf{Frame-Inference-via-Text}: Use the textual versions of the two neural DRGs described in the previous subsection. The DRGs are fine-tuned on the T5-base model.
    \item \textbf{Frame-Inference-via-Semantic-Parsing}: Use the semantic-parse versions of the two neural DRGs described in the previous section (fine-tuned on T5-base).
    \item \textbf{Frame-Inference-via-Semantic-Parsing-fine-tuned-Glucose}: Same as experiment 3 except the DRGs are fine-tuned on the Glucose rule generation model, instead of starting with T5-base.
\end{enumerate}

Experiments 2-4 use Braid to solve the problem. We initialize a knowledge-base consisting of the 4 sentence story, represented either as textual propositions (Expt. 2), or semantic parse propositions (Expt. 3 / 4).

Note that in experiments 3 and 4, using semantic parsing to map textual phrases to predicates (as shown in Figure \ref{fig:rocInference}) and using those predicate embeddings for unification is exactly our weak-unification approach, while the textual approach in experiment 2 maps phrases to syntactic parse predicates and uses exact (strict) unification for matching.

%For fine-tuning the neural models, we used Hugging Face's transformers library. 
All DRG models were trained on a single GPU with an effective batch size of \texttt{64}, a learning rate of \texttt{1e-3} and warmup=\texttt{100} steps. Models for frame-detection (step 1) were trained for \texttt{100} epochs, while models for ending-prediction (step 2) were trained for \texttt{10} epochs.

\begin{table}[htb]
    \centering
    \begin{tabular}{|p{7.5cm}|p{1cm}|}
        \hline
        \textbf{Model} & \textbf{Acc.} \\\hline
        E2E Neural Baseline & 86.15\% \\ \hline
        Braid: Frame Inf (Text) & 87.17\% \\ \hline
        Braid: Frame Inf (Sem Parse) & 87.76\% \\ \hline 
        \textbf{Braid: Frame Inf (Sem Parse + Glucose-tuned)} & \textbf{89.88\%} \\ \hline 
        HintNet \cite{HintNet} & 79.2\% \\ \hline
        GPT2 \cite{gpt2} & 86.5\% \\ \hline
        ISCK \cite{chenatal} & 87.6\% \\ \hline
        BERT-base + MNLI \cite{li2019story} & 90.6\% \\ \hline
    \end{tabular}
    \caption{ROC Story Cloze (Spring 2016) Test Results}
    \label{table:rocResults}
\end{table}

We use a two stage prover to solve the story cloze task: the prover first issues an open variable frame query (\texttt{frame(?X)}) to the frame-detection-DRG (which uses the story KB as context) to infer potential frames. These frame inferences are added to the KB via the generated rules. Then, the prover issues a query for each ending choice to the second DRG, which uses the story and frames as context to produce rules deriving the ending. Lastly, Braid searches for all proofs for each of the endings (in this case, each proof is a linear chain back from the ending to the frame and then to the story). The final proof scoring function marginalizes across all the intermediate frame propositions. 

Results of the experiments are shown in Table \ref{table:rocResults}. The table also includes the performance scores from other SOTA systems on this task, as a point of comparison. We can see that Braid is highly competitive with the SOTA systems while providing frame-based explanations (e.g. Figure \ref{fig:rocInference}). %Also, note that the frame inference experiments were done with \emph{bronze} frame labeled data, by manually annotating a small subset of the validation set, so there is clear headroom for improvement.  

%While the above solution does not use deep or complex logical reasoning, it shows how Braid can be customized for a simple rule-chaining task considering statistical rule generators and achieve close to SOTA results. %(and the custom prover and proof-scoring function makes it operate like a classical probabilistic model). 

\section{Conclusion}

We describe Braid, a novel reasoner that combines symbolic reasoning with statistical functions for fuzzy unification and dynamic rule generation. To manage the larger proof search space (due to the relaxed unification and the probabilistic-rules generated on the fly), our implementation uses a best-first search algorithm in a distributed task-based framework to construct proof graphs iteratively. Finally, we show how Braid is adapted for a commonsense reasoning task (ROC Story Cloze), and achieves close to SOTA results while generating logical explanations.

%% The file kr.bst is a bibliography style file for BibTeX 0.99c
\bibliography{braid, nonacl}

\begin{thebibliography}{14}
\providecommand{\natexlab}[1]{#1}

\bibitem[{Alec et~al.(2019)Alec, Jeff, Rewon, David, Dario, and Ilya}]{gpt2}
Alec, R.; Jeff, W.; Rewon, C.; David, L.; Dario, A.; and Ilya, S. 2019.
\newblock Language Models are Unsupervised Multitask Learners.
\newblock Technical report, OpenAI.

\bibitem[{Chen, Chen, and Yu(2019)}]{chenatal}
Chen, J.; Chen, J.; and Yu, Z. 2019.
\newblock Incorporating structured commonsense knowledge in story completion.
\newblock Technical report, AAAI.

\bibitem[{Dunietz et~al.(2020)Dunietz, Burnham, Bharadwaj, Rambow, Chu-Carroll,
  and Ferrucci}]{jesse2020}
Dunietz, J.; Burnham, G.; Bharadwaj, A.; Rambow, O.; Chu-Carroll, J.; and
  Ferrucci, D. 2020.
\newblock To Test Machine Comprehension, Start by Defining Comprehension.
\newblock In \emph{Proceedings of the 58th Annual Meeting of the Association
  for Computational Linguistics}, 7839--7859. Online: Association for
  Computational Linguistics.

\bibitem[{Kalyanpur et~al.(2020)Kalyanpur, Biran, Breloff, Chu-Carroll,
  Diertani, Rambow, and Sammons}]{spindle2020}
Kalyanpur, A.; Biran, O.; Breloff, T.; Chu-Carroll, J.; Diertani, A.; Rambow,
  O.; and Sammons, M. 2020.
\newblock Open-Domain Frame Semantic Parsing Using Transformers.
\newblock arXiv:2010.10998.

\bibitem[{Leon et~al.(2019)Leon, Pasquale, Jannes, Ulf, and Tim}]{Weber2019}
Leon, W.; Pasquale, M.; Jannes, M.; Ulf, L.; and Tim, R. 2019.
\newblock NLProlog: Reasoning with Weak Unification for Question Answering in
  Natural Language.
\newblock In \emph{Proceedings of the 57th Annual Meeting of the Association
  for Computational Linguistics}, 6151--6161. Florence, Italy: Association for
  Computational Linguistics.

\bibitem[{Li, Ding, and Liu(2019)}]{li2019story}
Li, Z.; Ding, X.; and Liu, T. 2019.
\newblock Story Ending Prediction by Transferable BERT.
\newblock arXiv:1905.07504.

\bibitem[{Mostafazadeh et~al.(2020)Mostafazadeh, Kalyanpur, Moon, Buchanan,
  Berkowitz, Biran, and Chu-Carroll}]{Glucose}
Mostafazadeh, N.; Kalyanpur, A.; Moon, L.; Buchanan, D.; Berkowitz, L.; Biran,
  O.; and Chu-Carroll, J. 2020.
\newblock GLUCOSE: GeneraLized and COntextualized Story Explanations.
\newblock In \emph{Proceedings of the 2020 Conference on Empirical Methods in
  Natural Language Processing}.

\bibitem[{Nasrin~Mostafazadeh(2016)}]{ROC}
Nasrin~Mostafazadeh, e.~a. 2016.
\newblock A Corpus and Cloze Evaluation for Deeper Understanding of Commonsense
  Stories.
\newblock In \emph{Proceedings of the 2016 Conference of the North American
  Chapter of the Association for Computational Linguistics: Human Language
  Technologies}, 839--849. San Diego, California: Association for Computational
  Linguistics.

\bibitem[{Riegel et~al.(2020)Riegel, Gray, Luus, Khan, Makondo, Akhalwaya,
  Qian, Fagin, Barahona, Sharma, Ikbal, Karanam, Neelam, Likhyani, and
  Srivastava}]{DBLP:IBM_LNN}
Riegel, R.; Gray, A.~G.; Luus, F. P.~S.; Khan, N.; Makondo, N.; Akhalwaya,
  I.~Y.; Qian, H.; Fagin, R.; Barahona, F.; Sharma, U.; Ikbal, S.; Karanam, H.;
  Neelam, S.; Likhyani, A.; and Srivastava, S.~K. 2020.
\newblock Logical Neural Networks.
\newblock \emph{CoRR}, abs/2006.13155.

\bibitem[{Ruppenhofer et~al.(2006)Ruppenhofer, Ellsworth, Petruck, Johnson, and
  Scheffczyk}]{framenet:book}
Ruppenhofer, J.; Ellsworth, M.; Petruck, M.~R.; Johnson, C.~R.; and Scheffczyk,
  J. 2006.
\newblock \emph{FrameNet II: Extended Theory and Practice}.
\newblock Berkeley, California: International Computer Science Institute.
\newblock Distributed with the FrameNet data.

\bibitem[{Schank and Abelson(1975)}]{scripts}
Schank, R.~C.; and Abelson, R.~P. 1975.
\newblock Scripts, plans, and knowledge.
\newblock \emph{Proceedings of the 4th international joint conference on
  Artificial intelligence}, 1: 151--157.

\bibitem[{Serafini and d'Avila Garcez(2016)}]{Serafini2018}
Serafini, L.; and d'Avila Garcez, A.~S. 2016.
\newblock Logic Tensor Networks: Deep Learning and Logical Reasoning from Data
  and Knowledge.
\newblock \emph{In Proceedings of AI*IA}, 334–348.

\bibitem[{Stevenson and Lindberg(2010)}]{noad}
Stevenson, A.; and Lindberg, C.~A., eds. 2010.
\newblock \emph{New Oxford American Dictionary (3 ed.)}.
\newblock Oxford University Press.
\newblock Electronic Version NOAD3A+2019\_G1.

\bibitem[{Zhou, Huang, and Zhu(2019)}]{HintNet}
Zhou, M.; Huang, M.; and Zhu, X. 2019.
\newblock Story ending selection by finding hints from pairwise candidate
  endings.
\newblock In \emph{TASLP}.

\end{thebibliography}

\end{document}